\documentclass[pmlr]{jmlr}

\usepackage{wrapfig}
\usepackage[normalem]{ulem}
\useunder{\uline}{\ul}{}
\usepackage{multirow}
\usepackage[framemethod=TikZ]{mdframed}
\usepackage[table,xcdraw]{xcolor}
\usepackage{mathtools, nccmath}

\definecolor{olivegreen}{rgb}{0.01, 0.75, 0.24}
\usepackage{longtable}
\usepackage[export]{adjustbox}
 \usepackage{booktabs}
 \usepackage{amssymb}
\usepackage{tcolorbox}
\usepackage{algorithm}
\usepackage{algorithmic}
\usepackage{threeparttable}
\usepackage{caption}
\usepackage{makecell}

 \DeclareMathOperator{\EX}{\mathbb{E}}
\usepackage[load-configurations=version-1]{siunitx} 

\makeatletter
\def\set@curr@file#1{\def\@curr@file{#1}} 
\makeatother

\theorembodyfont{\upshape}
\theoremheaderfont{\scshape}
\theorempostheader{:}
\theoremsep{\newline}

\jmlrvolume{219}
\jmlryear{2023}
\jmlrworkshop{Machine Learning for Healthcare}

\title[Deep Hazard Analysis]{Maximum Likelihood Estimation of Flexible Survival Densities with Importance Sampling}

\author{\Name{Mert Ketenci}
\Email{mk4139@columbia.edu}\\ 
\addr Department of Computer Science\\
Columbia University\\
New York, NY, USA
\AND
\Name{Shreyas Bhave}
\Email{sab2323@cumc.columbia.edu }\\ 
\addr Department of Biomedical Informatics\\
Columbia University\\
New York, NY,  
\AND
\Name{Noémie Elhadad}
\Email{noemie.elhadad@columbia.edu }\\ 
\addr Department of Biomedical Informatics\\
Columbia University\\
New York, NY, USA
\AND
\Name{Adler Perotte}
\Email{ajp2120@cumc.columbia.edu}\\ 
\addr Department of Biomedical Informatics\\
Columbia University\\
New York, NY,  } 

\begin{document}

\maketitle

\begin{abstract}

Survival analysis is a widely-used technique for analyzing time-to-event data in the presence of censoring. In recent years, numerous survival analysis methods have emerged which scale to large datasets and relax traditional assumptions such as proportional hazards. These models, while being performant, are very sensitive to model hyperparameters including: (1) number of bins and bin size for discrete models and (2) number of cluster assignments for mixture-based models. Each of these choices requires extensive tuning by practitioners to achieve optimal performance. In addition, we demonstrate in empirical studies that: (1) optimal bin size may drastically differ based on the metric of interest (e.g., concordance vs brier score), and (2) mixture models may suffer from mode collapse and numerical instability. We propose a survival analysis approach which eliminates the need to tune hyperparameters such as mixture assignments and bin sizes, reducing the burden on practitioners. We show that the proposed approach matches or outperforms baselines on several real-world datasets.

\end{abstract}

\section{Introduction}\label{sec:intro}

Survival analysis is concerned with modeling time-to-event data by estimating the probability that an event will occur at a future time. Time-to-event data differ from other data by censoring; for certain data points, the true event times may be unobserved. Data is often right censored, indicating that the event occurred after the censoring time but the exact time is unknown. In healthcare settings, survival analysis is useful in a wide variety of applications where censoring naturally occurs, including predicting the risk of disease and death \citep{vigano2000survival, perotte2015risk, nagpal2021dcm} and analyzing clinical trial data \citep{fleming2000survival, faucett2002survival}.  


Traditionally, survival analysis data has been analyzed using models such as Cox proportional hazards or by fitting the time-to-event distribution using a simple, unimodal parametric distribution, such as the Weibull distribution \citep{cox1972regression}.

In recent years, there have been numerous flexible survival analysis methods introduced which relax these assumptions and are performant on a number of real-world datasets \citep{ishwaran2008random,katzman2018deepsurv,kvamme2019time,nagpal2021dsm, nagpal2021dcm}. However, this flexibility has often come with the cost of introducing many additional hyperparameters. Thus, these models require extensive exploration of the hyperparameter space to achieve optimal performance, greatly elevating the burden on practitioners.


Existing survival methods that work with individual-level time-to-event predictions can be grouped into three main categories \citep{haider2020effective}: (1) parametric (2) semi-parametric, and (3) non-parametric.



Parametric survival methods assume a known probability distribution over time-to-event data, conditioned on certain covariates, and optimize the log-likelihood or the evidence-lower-bound (ELBO). Known probability distributions (e.g. Weibull, Log-Normal) have the shortcoming that they are constrained by particular hazard function shapes and are unimodal. Some methods have aimed to relax distributional assumptions by using discrete, categorical distributions \citep{lee2018deephit,miscouridou2018deep}, while others have taken the approach of using continuous, flexible distributions using mixture models \citep{nagpal2021dsm, han2022survival}. With the added flexibility, these models have been shown to outperform numerous baseline methods with more restrictive assumptions. However, they also introduce additional hyperparameters. For discrete models, there is the added challenge of specifying the appropriate number of bins and bin sizes. For mixture-based models, the number of mixture distributions must be specified and even with a sufficiently large number of mixtures, the models may collapse to local optima during training resulting in pathologies such as mode collapse \citep{shireman2016local, makansi2019overcoming}. In addition, these models can be numerically unstable \citep{makansi2019overcoming}.

Most commonly used semi-parametric methods, such as Cox proportional hazards (CoxPH) and DeepSurv, are constrained by the proportional hazards assumptions \citep{cox1972regression, katzman2018deepsurv}. Proportional hazards assume that different instances follow the same hazard trajectory, up to a multiplicative constant, which can be too strict for real-world time-to-event data. Recently, \cite{kvamme2019time} proposed extending CoxPH using a flexible non-proportional hazard function. However, their model does not allow for data sub-sampling and requires gradient approximations that are biased. 

Random Survival Forests (RSF) is a well-known non-parametric method for survival analysis \citep{ishwaran2008random}.
When tuned carefully, RSF performs on par with or better than the most recent state-of-the-art approaches. However, as also indicated by \citet{nagpal2021dcm}, RSF is sensitive to certain hyperparameters and require careful tuning. \\

In this paper, we introduce a flexible parametric survival analysis approach that directly models the hazard function to address the above gaps. Our contributions are as follows:

\begin{enumerate}
    \item We derive a continuous-time non-proportional survival model whose hazard function can take any shape, both over time and covariates.


    \item We introduce an importance sampling (IS) method for estimating the gradient of the otherwise intractable full log-likelihood. Our algorithm scales well to large datasets without requiring biased approximations such as sub-sampling risk sets, numerical integration \citep{butler1985statistical, kvamme2019time, danks2022derivative} and computationally expensive approaches that require ODE Solvers \citep{tang2022soden}. 
    
    \item To the best of our knowledge we are the first to propose an unbiased full log-likelihood optimization method for a non-proportional flexible hazard function without using a mixture model.

    \item Other than the network architecture, we only have a single hyperparameter which is the number of importance samples, but this hyperparameter is guaranteed to only improve estimates with larger sizes as we show empirically. 
    
    \item We carry out in-depth experimental analysis on several real-world datasets, empirically show the advantages of our approach, and demonstrate that it consistently performs well.
\end{enumerate}

\subsection*{Generalizable Insights about Machine Learning in the Context of Healthcare}

    Survival analysis has numerous applications in healthcare including risk assessment for chronic diseases and analysis of clinical trial data. In recent years, many methods have emerged which scale to large datasets and relax the restrictive assumptions of widely-used approaches such as Cox proportional hazards. However, this added flexibility has come at the cost of introducing many hyperparameters (e.g., bin size, number of bins, number of mixture components) as well as optimization challenges. This increases the burden placed on practitioners to explore the large space of hyperparameters to ensure optimal performance. We sought to eliminate the necessity for these hyperparameters, while also maintaining all the favorable properties of these models (i.e., flexible, scalable, unbiased estimates, optimized via stochastic gradient descent, continuous). Our model is run with default parameters on all datasets and is able to match or outperform existing state of the art methods. We believe this model will ease the burden on practitioners for fitting new datasets. We have made the code publicly available on GitHub with instructions on how to fit our model on any dataset quickly and efficiently. 



\section{Sensitivity of Existing Methods to Hyperparameters}

\begin{figure}[h]
\centering
\includegraphics[width=\linewidth]{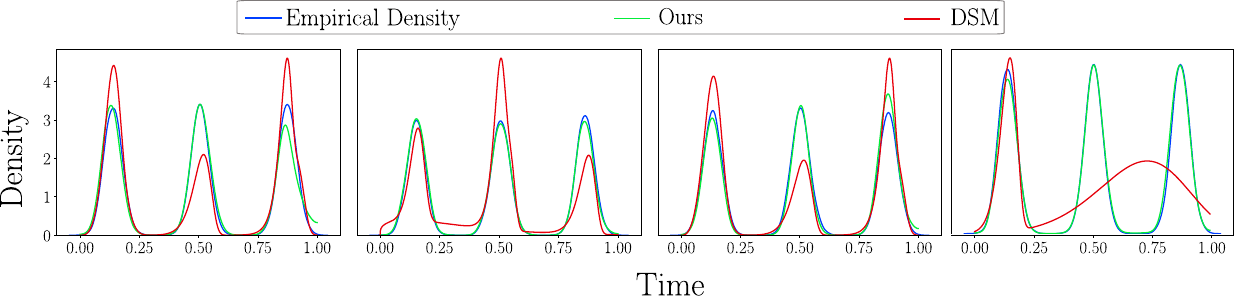}
\caption{Illustration of our model's ability to capture multi-modal survival densities without requiring a mixture density like Deep Survival Machines (DSM), Deep Cox Mixtures (DCM), and SurvMDN \citep{nagpal2021dsm, nagpal2021dcm, han2022survival}. We simuate a conditional time-to-event dataset using the generative story $t \sim \sum_{j=1}^3 \frac{1}{3} \mathcal{N}(\mu_j + \boldsymbol{x}_i, 1) \quad \text{and} \quad \boldsymbol{x}_i \sim \mathcal{N}(0, 0.01)$ with $\mu_j = 10j$ (we scale the data in $(0, 1]$ to ensure positivity) and fit our model and DSM (with 5 components) over 4 different runs using full log-likelihood optimization. We plot the time-to-event density of a random instance. We observe that the mixture density network can settle in a local-optimum solutions easier. On the other hand, we have a more stable estimation approach which captures the density well across 4 random runs.} 
\label{fig:mixture}
\end{figure}

One class of continuous time models that can approximate any distribution is mixture density networks. A large number of density components, in theory, can represent any distribution \citep{bishop2006pattern}. Examples in survival analysis include \citet{nagpal2021dsm, nagpal2021dcm, han2022survival}. However, mixture models are prone to arrive at locally optimal solutions resulting in mode collapse and poor density estimation \citep{shireman2016local, makansi2019overcoming}. We further demonstrate this empirically by running simulation studies, the details of which are specified in Figure \ref{fig:mixture}.

Another important class of survival models discretize time with bins for flexible density estimation \citep{ranganath2016deep, miscouridou2018deep, lee2018deephit}. However, this approach introduces an important hyperparameter: the number of bins. To study the sensitivity of these models to this hyperparameter, we design a simple experiment using DeepHit \citep{lee2018deephit} as the model and the commonly used SUPPORT dataset \citep{knaus1995support} for benchmarking survival models. 

\begin{figure}[h]
\centering
\includegraphics[width=\linewidth]{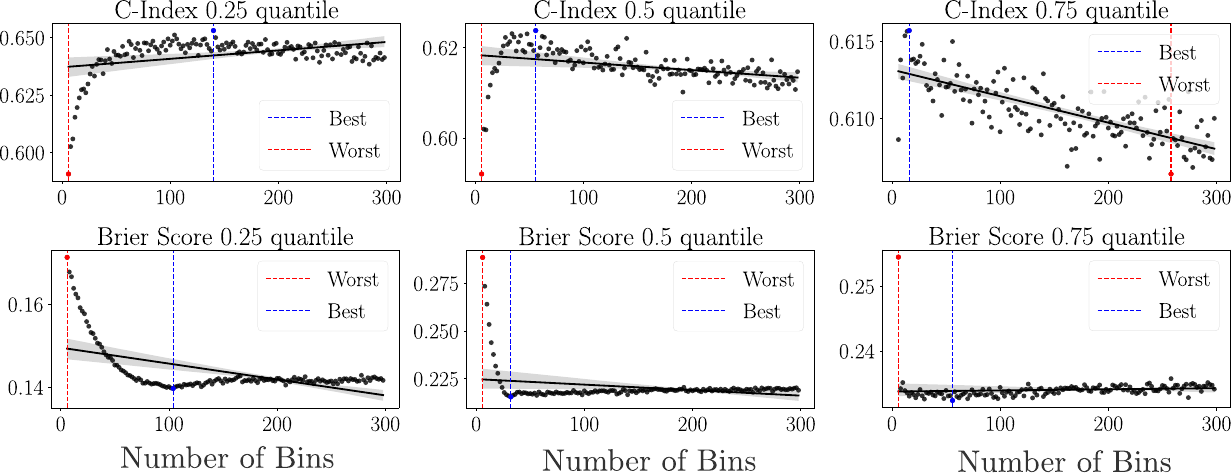}
\caption{The sensitivity of discrete time models to the number of bins. The \color{blue} blue \color{black} and  \color{red} red \color{black} lines describe the best and worst bin sizes for each metric. We employ DeepHit for training and run 750 different runs using 5 fold cross-validation and 150 bin sizes (ranging from 2 to 300 with equal spacing). }
\label{fig:bins}
\end{figure}

Figure \ref{fig:bins} shows results averaged across held-out folds for both concordance and brier score as a function of the number of bins. For both metrics, there is an optimal bin number with the metrics performing worse when specifying both a small and very large number of bins. The optimal number of bins can vary significantly across metrics, making it difficult to specify a single number. This typically results in a trade-off between model calibration and ranking.


\section{Deep Hazard Analysis}
In this section, we describe our approach called Deep Hazard Analysis (DHA). First, we describe our model in detail in Section \ref{sec:model}. Then we describe parameter estimation in Section \ref{sec:est} and how to make predictions in Section \ref{sec:pred}. Finally, we describe our network architecture in Section \ref{sec:arc}.
\subsection{Model}\label{sec:model}

Let $\{\boldsymbol{x}_i, t_i, \delta_i\}_{i=1}^N$ be a right-censored time-to-event data where $\boldsymbol{x}_i\in \mathbb{R}^d$, $t_i\in \mathbb{R}^+$ and $\delta_i \in \{0,1\}$ stand for covariates, time-to-event, and censoring indicator, respectively. Then we write our probability density function, using the hazard and survival functions, as:

\begin{align}
    f(\boldsymbol{x}, t;\theta) &= \underbrace{\lambda(\boldsymbol{x}, t;\theta)}_{\text{Hazard}} \underbrace{S(\boldsymbol{x},t;\theta)}_{\text{Survival}}\\
    &= \lambda(\boldsymbol{x},t;\theta) \exp \{-\Lambda(\boldsymbol{x},t;\theta)\}\\
    &= \lambda(\boldsymbol{x},t;\theta) \exp \left\{-\int_0^t \lambda(\boldsymbol{x},t; \theta) dt.\right\}.
\end{align}

Here, $ \lambda(\boldsymbol{x},t, \theta) = \log\left(1 + \exp \{ \Phi(\boldsymbol{x}, t;\theta)\}\right)$, where $\Phi(.;\theta)$ is a flexible function approximator. In this paper, we use neural networks to model $\Phi(.;\theta)$. The log-likelihood of this survival model for $N$ data points is:

\begin{align}
     \ell &=\sum_{i=1}^N \delta_i \log \lambda(\boldsymbol{x}_i, t_i;\theta) + \log S(\boldsymbol{x}_i,t_i;\theta) dt \\
    &= \sum_{i=1}^N \left(\delta_i \log \lambda(\boldsymbol{x}_i,t_i; \theta) - \int_0^{t_i} \lambda(\boldsymbol{x}_i,t; \theta) dt \right), \label{eq:ll}
\end{align}

Note that, the hazard function implied by our model is not restricted by the proportional hazards assumption.

\subsection{Parameter Estimation}\label{sec:est}

The integral in Equation \ref{eq:ll} is intractable. A straight-forward approach to approximate it in an unbiased way is by importance sampling:

\begin{align}
    \ell&=\sum_{i=1}^N \left(\delta_i \log \lambda(\boldsymbol{x}_i,t_i; \theta) - \int_0^{t_i} \lambda(\boldsymbol{x}_i,t; \theta) dt \right)\\
    &= \sum_{i=1}^N\left(\delta_i \log \lambda(\boldsymbol{x}_i,t_i; \theta) - t_i \int_0^{t_i} \frac{\lambda(\boldsymbol{x}_i,t; \theta)}{t_i} dt \right) \label{eq:int} \\
    &= \sum_{i=1}^N\left( \delta_i \log \lambda(\boldsymbol{x}_i,t_i; \theta) - t_i \EX_{t\sim U(0, t_i)} \left[ \lambda(\boldsymbol{x}_i,t; \theta) \right] \right) \label{eq:ex} \\
    &= N \EX_{\mathcal{D}} \left[\delta_i \log \lambda(\boldsymbol{x}_i,t_i; \theta) - t_i \EX_{U(0, t_i)} \left[ \lambda(\boldsymbol{x}_i,t; \theta) \right] \right]. \label{eq:loss}
\end{align}

Here $U(.)$ is the uniform distribution. Note that the integral in Equation \ref{eq:int} is the expected value of the hazard w.r.t. time, on a uniform grid, allowing for Equation \ref{eq:ex}. We approximate Equation \ref{eq:loss} by drawing $L$ Monte-Carlo samples from the empirical data distribution $\mathcal{D}$ and a set $\tilde{\boldsymbol{T}}_i = \{\tilde{t}_{ij}\}_{j=1}^M$ from the IS distribution $U(0, t_i)$. The loss function after using a mini-batch of $L$ data and $M$ IS samples is denoted as:

\begin{align}
\begin{split}\label{eq:mcloss}
    &\tilde{\ell} (\boldsymbol{T}|\boldsymbol{X}, \Delta, \boldsymbol{\tilde{T}}; \theta)\\
    &= \frac{N}{L} \sum_{i=1}^L \left(\delta_i \log \lambda(\boldsymbol{x}_i,t_i; \theta)  - \frac{t_i}{M} \sum_{j=1}^M \lambda(\boldsymbol{x}_i,\tilde{t}_{ij} ; \theta)  \right),
\end{split}
\end{align}

where $\boldsymbol{T} = \{t_i\}_{i=1}^L$, $\boldsymbol{X} = \{\boldsymbol{x}_i\}_{i=1}^L$, $\Delta = \{\delta_i\}_{i=1}^L$, and $\tilde{\boldsymbol{T}} = \{\tilde{\boldsymbol{T}}_i\}_{i=1}^L$. We describe our learning algorithm in Algorithm \ref{alg:lrn}. 

\begin{algorithm}[h]
    \caption{Mini-batch stochastic gradient descent algorithm for our model.}\label{alg:lrn}
\begin{algorithmic}
    \STATE {\bfseries Input:}  $\mathcal{D}$
    \STATE $\theta \leftarrow{}$ Initialize parameters
    \WHILE{not converged}
    \STATE  $\{\boldsymbol{X}, \boldsymbol{T}, \Delta \}\leftarrow{}$ Sample $L$ data from $\mathcal{D}$
    
    \STATE $\tilde{\boldsymbol{T}} \leftarrow{}$ Sample $M$ importance samples from $U(0, \boldsymbol{T})$
  
    \STATE   ${g}$ $\leftarrow{}$ $\nabla_{\theta }\tilde{\ell} (\boldsymbol{T}|\boldsymbol{X}, \Delta, \boldsymbol{\tilde{T}}; \theta)$
 
    \STATE $\theta$ $\leftarrow{}$ Update using gradients ${g}$ 

   \ENDWHILE
   \STATE {\bfseries Output:}  $\theta$
\end{algorithmic}
\end{algorithm}

\begin{figure}[!htb]
    \raggedleft
    \begin{minipage}{0.53\textwidth}
        \raggedleft
        \includegraphics[width=0.9\linewidth, right]{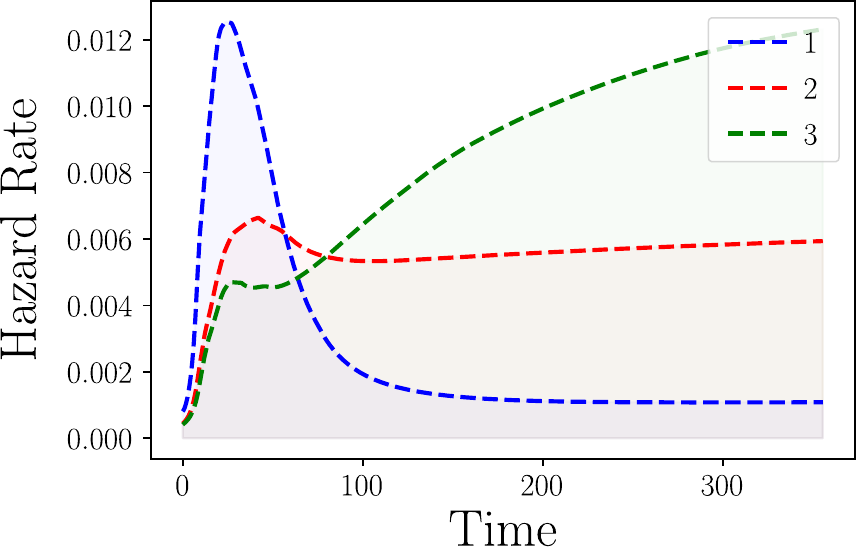}
        \centering
        \caption{Illustration of our model's ability to capture non-proportional hazard functions on 3 random instances from the METABRIC dataset. \color{blue}Blue\color{black}, \color{olivegreen}green\color{black}, and \color{red}red \color{black} curves denote different patients. Unlike proportional models, such as CoxPH and DeepSurv, the hazard rate and its shape differ between patients. Patient 1 is under high instantaneous risk in the early stages, which later decreases significantly. The instantaneous risk for patient 3 increases by time while patient 2 remains relatively constant. Our model does not need for critical hyperparameters such as cluster size, discretization or ODESolvers to model non-linear hazard rates.}
        \label{fig:hazard}
    \end{minipage}\hfil
    \begin{minipage}{0.4\textwidth}
Importance sampling can be problematic in high-dimensional spaces with an exponential growth in variance \citep{scharth2016particle}. However, we are strictly interested in modeling 1-dimensional time-to-event data. We demonstrate the stability and convergence of our learning process in Figure \ref{fig:lrn}. Moreover, the support of the uniform distribution is well-defined over $0$ and $t_i$ eliminating the need for rejecting any sample.
\paragraph{Unbiased gradients.} An important property that distinguishes our approach from \citet{kvamme2019time} is that Equation \ref{eq:mcloss} allows for unbiased learning of $\theta$. Hence, $\nabla_\theta \tilde{\ell} (\boldsymbol{T}|\boldsymbol{X}, \Delta, \boldsymbol{\tilde{T}}; \theta)$ is an unbiased Monte-Carlo estimate of the true gradients, $\nabla_\theta\ell$:
    \end{minipage}
\end{figure}

\begin{figure}[!htb]
    \centering
    \begin{minipage}{.4\textwidth}
\subsection{Predictions}\label{sec:pred}

The quantity of interest is the probability of survival of an instance above some point in future $t_i$ denoted by $S(t_i | \boldsymbol{x}_i; \theta)$, which is analytically intractable. The IS method can be employed here to predict the survival as:

\begin{align}
      &S(t_i , \boldsymbol{x}_i; \theta) \\
      &= \medmath{\exp \left\{-\int_0^{t_i}\lambda(\boldsymbol{x}_i,t; \theta)  dt\right\}} \\
     &=  \exp \left\{- t_i \EX_{U(0,t_i)}\left[ \lambda(\boldsymbol{x}_i,t; \theta)\right]\right\}\\
     &\approx \exp \left\{-  \frac{t_i}{M} \sum_{j=1}^M \lambda(\boldsymbol{x}_i,\tilde{t}_{ij} ; \theta) \right\}\\
     &= \tilde{S}(t_i , \boldsymbol{x}_i, \tilde{\boldsymbol{T}}_i; \theta),\label{eq:pred}
\end{align}

This is important because it guarantees that the model parameters $\theta$ converge to a value that optimizes the full log-likelihood $\ell$ while scaling to large datasets by data sub-sampling.

    \end{minipage}\hfil
    \begin{minipage}{0.5\textwidth}
        \centering
        \includegraphics[width=\linewidth, right]{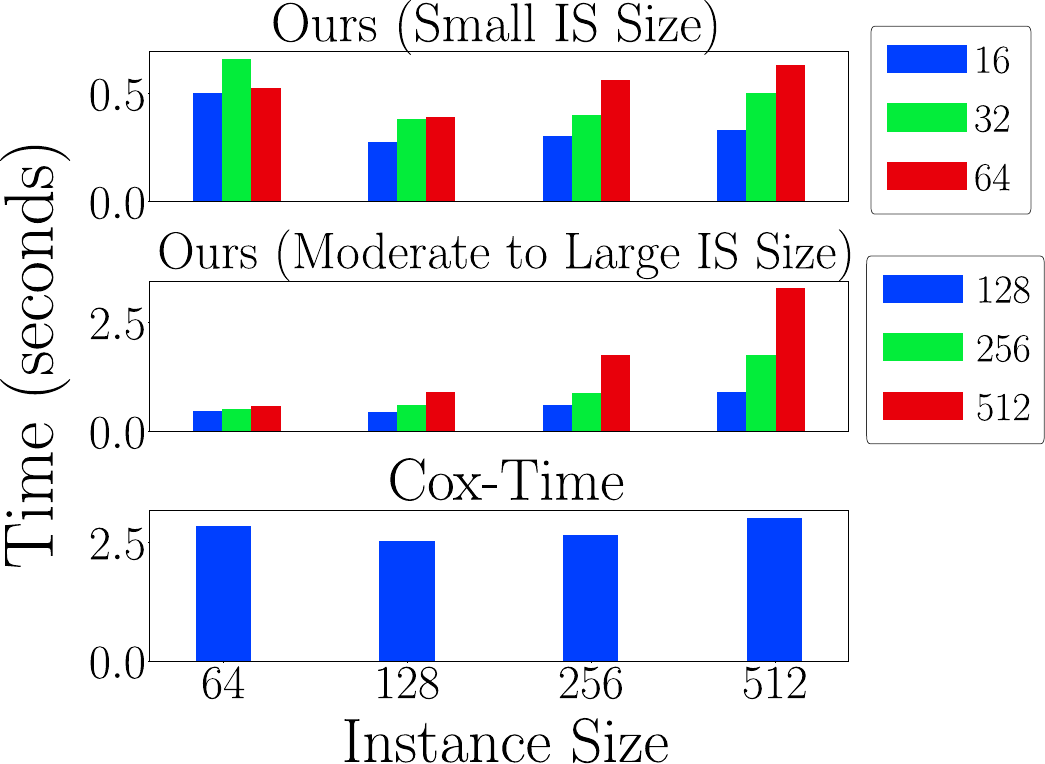}
        \caption{The time required to predict different IS and patient sizes on the METABRIC dataset. \color{blue}Blue\color{black}, \color{olivegreen}green\color{black}, and \color{red}red \color{black} bars denote the IS sizes. Empirically, we observe that prediction time scales linearly with respect to IS and instance sizes  for moderate to large IS sizes. This result is as expected. For IS sizes of 16, 32, 48, and 64 altering the instance size does not result in a  major difference in computational time as discussed in Section \ref{sec:pred}. We also show the Cox-Time model to provide a baseline for the prediction times. We did not add the time required to fit the baseline hazard function for the Cox-Time model. Our implementation is approximately 5 times faster than the Cox-Time model for IS size $\leq$ 64. For a fair comparison, we use the same neuron, layer sizes, and GPU for both models. The numbers are reported for single-precision floating-point format.}
        \label{fig:time}
    \end{minipage}
\end{figure}

\begin{align}
\EX_{\mathcal{D}} \left[\nabla_\theta \tilde{\ell}  (\boldsymbol{T}|\boldsymbol{X}, \Delta, \boldsymbol{\tilde{T}}; \theta)\right]
&=\EX_{\mathcal{D}} \Bigg[ \nabla_\theta\frac{N}{L}\sum_{i=1}^L  \Bigg( \delta_i \log \lambda(\boldsymbol{x}_i,t_i; \theta) - \frac{t_i}{M} \sum_{j=1}^M \lambda(\boldsymbol{x}_i,\tilde{t}_{ij} ; \theta) \Bigg) \Bigg]\\
&=\nabla_\theta\frac{N}{L} \sum_{i=1}^L \Bigg(\EX_{\mathcal{D}} \Big[ \delta_i \log\lambda(\boldsymbol{x}_i,t_i; \theta) - \frac{t_i}{M} \sum_{j=1}^M \lambda(\boldsymbol{x}_i,\tilde{t}_{ij} ; \theta) \Big] \Bigg)\\
&=\nabla_\theta \frac{N}{L} L\EX_{\mathcal{D}} \left[\delta_i \log \lambda(\boldsymbol{x}_i,t_i; \theta) - \frac{1}{M} \sum_{j=1}^M  t_i\lambda(\boldsymbol{x}_i,\tilde{t}_{ij} ; \theta) \right]\\
&=\nabla_\theta  \underbrace{N \EX_{\mathcal{D}} \left[\delta_i \log \lambda(\boldsymbol{x}_i,t_i; \theta) -  t_i\EX_{U(0, t_i)} \left[   \lambda(\boldsymbol{x}_i,\tilde{t}_{ij} ; \theta) \right] \right] }_{\ell} \\
&= \nabla_\theta \ell.
\end{align}

\begin{figure}[h]
\centering
\includegraphics[width=\linewidth]{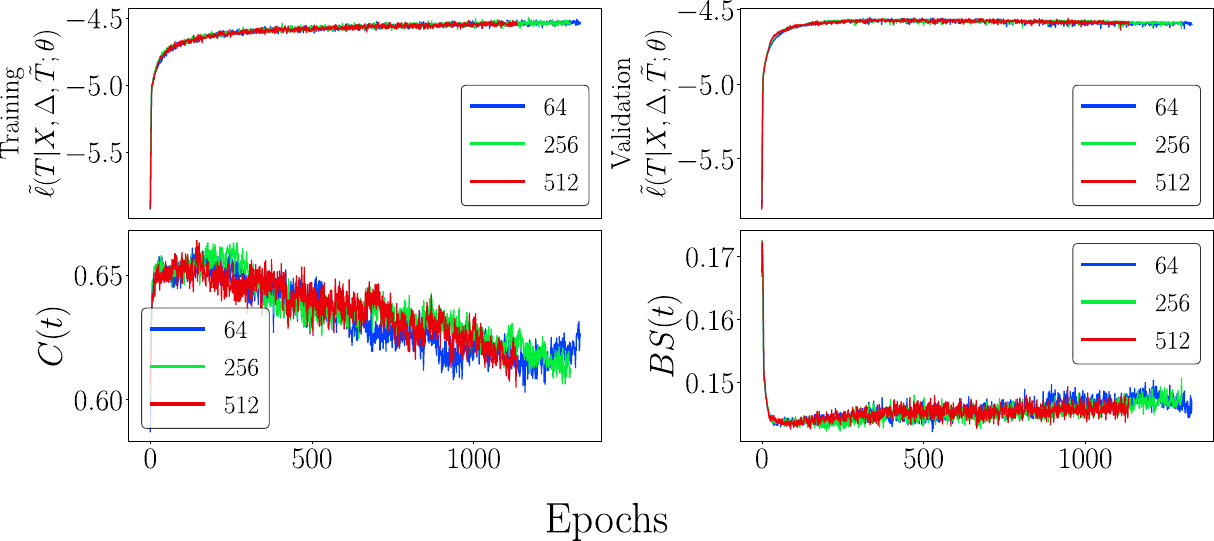}
\caption{Training and validation log-likelihood, C-Index ($C(t)$) and Brier Score ($BS(t)$), for $t : p(T < t) \leq 1/4$ (25$^\text{th}$ quantile), by epochs of our approach for different IS sizes. \color{blue}Blue\color{black}, \color{olivegreen}green\color{black}, and \color{red}red \color{black} lines denote the IS sizes. Training and validation log-likelihoods increase monotonically. Training stops when the validation log-likelihood does not improve for 800 epochs. Interestingly, lower IS sizes do not add a substantial variance and change the values the algorithm converges, for both training and validation log-likelihood values. The survival metrics for different IS sizes are also similar across different datasets, which empirically shows our algorithm's stability. We show this empirically for two different network architectures over three different importance sampling sizes on 4 real-world datasets, empirically. The empirical results are discussed in Section \ref{sec:res}.}
\label{fig:lrn}
\end{figure}

which requires evaluating a set $\tilde{\boldsymbol{T}}_i$ of $M$ importance samples with linear $\mathcal{O}(M)$ time complexity as shown in Algorithm \ref{alg:pred}. 

\begin{algorithm}[h]
    \caption{Making predictions.}\label{alg:pred}
\begin{algorithmic}
    \STATE {\bfseries Input:}  $\boldsymbol{x}_i, t_i, \theta$
    \STATE $\tilde{\boldsymbol{T}}_i \leftarrow{}$ Sample $M$ importance samples from $U(0, t_i)$
    \STATE $\tilde{\Lambda}(\boldsymbol{x}_i, \tilde{\boldsymbol{T}}_i;\theta) \leftarrow{}$ Calculate $  \frac{t_i}{M} \sum_{j=1}^M \lambda(\boldsymbol{x}_i,\tilde{t}_{ij} ; \theta) $
    \STATE $\tilde{S}(t_i , \boldsymbol{x}_i, \tilde{\boldsymbol{T}}_i; \theta)\leftarrow{} \exp \left\{- \tilde{\Lambda}(\boldsymbol{x}_i, \tilde{\boldsymbol{T}}_i;\theta) \right\}$ 
   \STATE {\bfseries Output:}  $\tilde{S}(t_i , \boldsymbol{x}_i, \tilde{\boldsymbol{T}}_i; \theta)$
\end{algorithmic}
\end{algorithm}

The time complexity for a set of $N$ instances can be thought of $\mathcal{O}(NM)$. However, GPUs allow for parallel computing over instances which practically results in $\mathcal{O}(M)$ operations for small IS sizes which  empirically show this in Figure \ref{fig:time}. $ \tilde{S}(t_i , \boldsymbol{x}_i, \tilde{\boldsymbol{T}}_i; \theta)\overset{p}{\to} S(t_i , \boldsymbol{x}_i; \theta)$ as $M$ increases, therefore it is beneficial to work with a relatively large M. We study the implications of IS to predictions empirically in  Section \ref{sec:eval}. 

\subsection{Network Architecture}\label{sec:arc}
We experiment with two neural network architectures to parameterize our model. Architecture 1 (A1) is formulated by:

\begin{align}
\begin{split}
\Phi(\boldsymbol{x},t;\theta) = \Phi_\text{shared}(\mathrm{cat}[\boldsymbol{x}, t];\theta_{\text{shared}}).
\end{split}
\end{align}

and (A2) is formulated by:

\begin{align}
\begin{split}
&\Phi(\boldsymbol{x},t;\theta)\\
&= \Phi_\text{shared}(\mathrm{cat}[\Phi_\text{cov}((\boldsymbol{x};\theta_{\text{cov}}), \Phi_\text{time}(t;\theta_{\text{time}})];\theta_{\text{shared}}).
\end{split}
\end{align}

\begin{figure}[h]
\centering
\includegraphics[width=\linewidth]{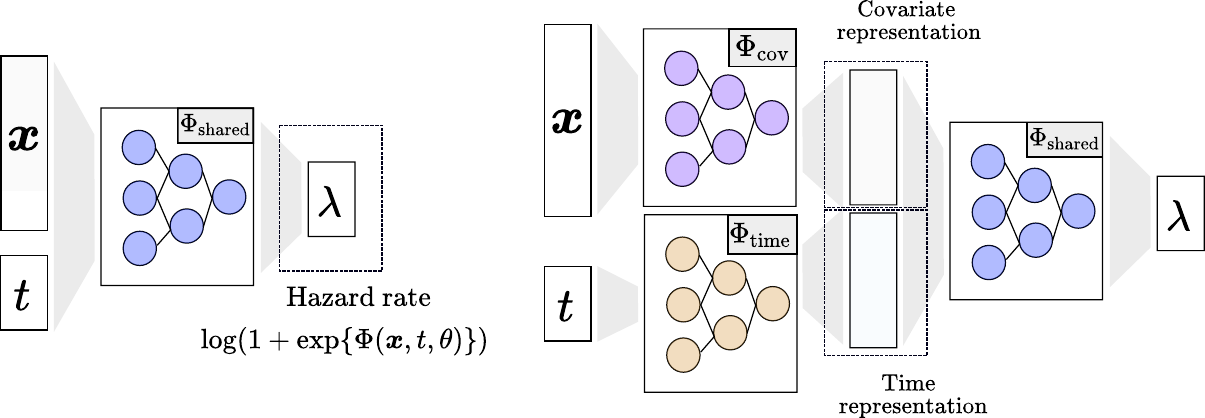}
\caption{The network architectures A1 (left) and A2 (right). A1 is a simple feed forward network parameterized by $\theta_{\text{shared}}$. A2 has two separate input layers for covariates $\boldsymbol{x}$ and time $t$. These layers form a latent representation of $\boldsymbol{x}$ and $t$ that are concatenated and fed into a shared layer followed by softplus function to model the hazard rate, $\lambda$.}
\label{fig:network}
\end{figure}

Here $\mathrm{cat}$ operation refers to concatenation of two vectors. 
Later, the output of the neural networks are feed into the softplus function to predict the hazard rate. The intuition behind the second architecture is to allow for learning a temporal embedding independent of patient covariates. The architectures are shown in Figure \ref{fig:network}.

\section{Related Work}

\paragraph{Parametric methods.} Deep Survival Analysis (DSA) and Deep Survival Machines (DSM) are two important examples of parametric methods \citep{ranganath2015survival,nagpal2021dsm}. DSA models time-to-event data using Weibull distribution conditioned on a latent representation drawn from a deep exponential family. A limitation of this approach is that it assumes proportional hazards with a Weibull base hazard rate for a fixed shape parameter. DSM models time-to-event data using a mixture of Log-normal and Weibull densities whose parameters are conditioned on individual instances and optimize the ELBO. Although a mixture of such densities allows for a more flexible hazard, the mixture size introduces an additional hyperparameter to tune.

Another important parametric line of work focuses on discretizing time. DeepHit is a well-known approach in this line of work \citep{lee2018deephit, lee2019dynamic}. This approach divides continuous time-to-event data into bins and assumes a categorical distribution. A limitation of this line of work is that it is sensitive to the heuristically chosen bin size. Additionally, it only allows for making predictions over a pre-defined set of time bins which can be problematic for instances that do not fall into this range requiring additional interpolations.

\paragraph{Semi-parametric methods.} The Cox proportional hazards model (CoxPH) is commonly employed in various settings due to its simplicity \citep{cox1972regression}. A significant amount of work focused on improving CoxPH. \citet{rosen1999mixtures} improved CoxPH using a mixture of linear experts. \citet{nagpal2019nonlinear} extended this approach using a variational objective. 

Other approaches that extend CoxPH involve using flexible function approximators. The Faraggi-Simon network was the first to couple CoxPH with neural networks \citep{faraggi1995neural}. However, this attempt did not demonstrate improvements over CoxPH \citep{mariani1997prognostic,xiang2000comparison,sargent2001comparison}. Later, \citet{katzman2018deepsurv} showed that modern deep-learning techniques improve CoxPH. A shortcoming of these approaches is the proportional hazards assumption, which can be too restrictive for real-world datasets.

More recently, Deep Cox Mixtures (DCM) proposed to use a mixture of non-linear Cox experts with the EM algorithm and further improved upon DeepSurv \citet{nagpal2021dcm}. There have also been methods that extended Cox's framework to handle unstructured data, such as images \citep{zhu2016deep, zhu2017wsisa, christ2017survivalnet}. 

\paragraph{Non-parametric methods.} RSF is an important method in this line of work \citep{ishwaran2008random}. RSF aggregates multiple trees by bagging and averages the result of each tree when making predictions. A limitation of the RSF is that the current implementations do not support GPUs, and it is unclear how to deploy them for datasets with large examples and covariates.

Amongst recent work, Cox-Time is the closest approach to DHA \citep{kvamme2019time}. Cox-Time extends CoxPH by removing the proportional hazards assumption and leveraging neural networks. Cox-Time is a semi-parametric model that optimizes partial log-likelihood. In its original form, Cox-Time is amenable to stochastic optimization and does not scale well to large datasets. Therefore, learning Cox-Time requires a biased but computationally cheaper gradient approximation. Unlike Cox-Time, our model (1) is parametric, (2) optimizes the full log-likelihood, and (3) does not require biased gradient approximations.

\section{Experiments}
In this section, we describe the datasets (\ref{sec:data}), baseline models (\ref{sec:base}) and evaluation metrics (\ref{sec:eval}) used to asses the performance of our model. We compare our model to multiple state-of-the-art approaches on three commonly used real-world datasets. We also investigate the impact of different IS sizes on these datasets.

\subsection{Datasets}\label{sec:data}
\paragraph{SUPPORT.} Study to Understand Prognoses Preferences Outcomes and Risks of Treatment (SUPPORT) consists of seriously ill hospitalized adult patients \citep{knaus1995support}. After preprocessing, there are 8873 instances and 23 covariates with median follow-up days and censoring rate of 231 and 31.9\%, respectively. We use the preprocessing of the PyCox library \citep{havakvpy36:online}.
\paragraph{METABRIC.} A Canada-UK project, the Molecular Taxonomy of Breast Cancer International Consortium (METABRIC) database, comprises targeted sequencing and survival information from breast cancer patients \citep{curtis2012genomic, pereira2016somatic}. After preprocessing, the dataset contains 1904 instances and 9 covariates with median follow-up days and a censoring rate of 114.9 days and 42\%, respectively. We use the preprocessing of the PyCox library \citep{havakvpy36:online}.
\paragraph{FLCHAIN.} A controlled trial conducted in Olmsted County, Minnesota that investigates the association of mortality and assay of serum free light chain (FLCHAIN) \citep{dispenzieri2012use}. After preprocessing, there are 6524 instances with 16 covariates. The median follow-up days and a censoring rate of 4303 days and 70\%, respectively. We use the preprocessing of the PyCox library \citep{havakvpy36:online}.
\paragraph{CKD/AKI.}
The study cohort consisted of 10,173 patients that are identified in chronic kidney disease (CKD) incident cohort where the event is defined as acute kidney injury (AKI) diagnosis, during hospitalization, by the Health Equity Research Assessment (HERA) characterization. The median follow-up days and censoring rate are 67 days and 64\%, respectively. 

\subsection{Baseline Models}\label{sec:base}

\paragraph{Deep Survival Machines (DSM).} A parametric survival model that extends beyond the AFT using a mixture of Weibull and log-normal distributions. The mixture assignments and time-to-event distributions are parameterized by neural networks conditioned on covariates. Parameter estimation is done by optimizing the ELBO, where the expectation is taken with respect to the conditional model prior \citep{nagpal2021dsm}. 

\paragraph{Deep Cox Mixtures (DCM).} A semi-parametric survival model that extends DeepSurv using a mixture of CoxPH components parameterized by neural networks. Parameter estimation is done by the EM algorithm and fitting polynomial splines to baseline hazards \citep{nagpal2021dcm}. 

\paragraph{DeepSurv.} A semi-parametric model that extends CoxPH by modeling relative risk using neural networks conditioned on covariates formulating a non-linear proportional hazard function\citep{katzman2018deepsurv}.

\paragraph{DeepHit.} A discrete-time survival model parameterized by neural networks with a softmax output layer. DeepHit uses cross-entropy loss combined with a ranking loss \citep{lee2018deephit}. 

\paragraph{Random Survival Forest (RSF).} An extension of random forests that fits multiple trees to survival data by bagging and using the cumulative hazard function computed by the Nelson-Aalen estimator \citep{ishwaran2008random}. 

\paragraph{Cox-Time.} A semi-parametric method that extends CoxPH beyond propotional hazards. Cox-Time uses neural networks to parameterize the hazard function \citep{kvamme2019time}. Parameter estimation is done by optimizing a biased approximation of the partial log-likelihood.

\paragraph{CoxPH.} Well-known semi-parametric Cox proportional hazards model \citep{cox1972regression}. Parameter learning is done by optimizing the partial log-likelihood.

\subsection{Evaluation Metrics}\label{sec:eval}

The literature on evaluation metrics for survival analysis is vast and beyond this paper's scope. We refer the reader to references in the following paragraphs for more detail. This paper focuses on evaluations over a fixed follow-up period. Fixed follow-up periods are used in many important real-world settings, such as randomized clinical trials. Similar to \citet{li2023spatio, nagpal2022counterfactual,  nagpal2021dsm, nagpal2021dcm, jeanselme2022neural,wang2022survtrace, lee2019temporal}, we consider event quantiles as follow-up periods and report concordance index (C-Index), Brier score (BS), and Area Under Receiver Operating Curve (ROC-AUC) metrics which are implemented by \citet{sksurv}. This provides an overview of how each model performs over time and helps ensure that the models effectively capture potential differences in risks over the event horizon. All metrics are adjusted by the inverse probability of censoring weight (IPCW), using the Kaplan-Meier censoring estimate, to account for the censoring bias \citep{kaplan1958nonparametric}.

\begin{table}[t]
\center
\begin{adjustbox}{width=\textwidth}
\begin{tabular}{c|ccc|ccc|ccc}
\toprule
\multicolumn{10}{c}{$\;\;\;\;\;\;\;\;\;\;\;\;\;\;\;\;$SUPPORT Dataset}                                                                                          \\ \midrule
                         & \multicolumn{3}{c}{25$^{\text{th}}$   Quantile}              & \multicolumn{3}{|c|}{50$^{\text{th}}$   Quantile}               & \multicolumn{3}{c}{75$^{\text{th}}$  Quantile}             
                         \\
                         \cline{2-10} 
\multirow{-2}{*}{Models} & C-Index        & BS             & ROC            & C-Index        & BS             & ROC            & C-Index        & BS             & ROC            \\ \hline
CoxPH                 & 0.553                & 0.262                & 0.558                & 0.567                & 0.222                & 0.587                & 0.590                & 0.351                & 0.649                \\
DeepSurv              & 0.603                & 0.143                & 0.607                & 0.598                & 0.217                & 0.618                & 0.610                & {\ul \textbf{0.231}} & \textbf{0.661}       \\
RSF                   & \textbf{0.657}       & \textbf{0.140}       & \textbf{0.664}       & 0.621                & 0.215                & 0.638                & \textbf{0.613}       & \textbf{0.232}       & {\ul \textbf{0.666}} \\
DeepHit               & 0.631                & 0.153                & 0.638                & 0.608                & 0.243                & 0.628                & 0.607                & 0.236                & \textbf{0.659}       \\
Cox-Time              & 0.640                & \textbf{0.140}       & 0.647                & 0.620                & 0.213                & 0.640                & {\ul \textbf{0.615}} & {\ul \textbf{0.231}} & \textbf{0.663}       \\
DSM                   & 0.645                & \textbf{0.140}                & 0.650                & 0.621                & 0.214                & 0.637                & \textbf{0.615}                & 0.245                & \textbf{0.657}                \\
DCM                   & \textbf{0.649}       & \textbf{0.140}       & \textbf{0.656}       & 0.616                & 0.216                & 0.633                & 0.602                & 0.236                & 0.650                \\
\Xhline{0.1\arrayrulewidth}
\rowcolor[HTML]{EFEFEF} 
DHA (IS =   64, A1) $\;$  & \textbf{0.651}       & {\ul \textbf{0.139}} & \textbf{0.660}       & 0.623                & 0.213                & 0.640                & \textbf{0.614}       & 0.236                & 0.644                \\
\rowcolor[HTML]{EFEFEF} 
DHA (IS =   256, A1) & \textbf{0.652}       & \textbf{0.140}       & \textbf{0.660}       & 0.624                & 0.213                & 0.641                & \textbf{0.614}       & 0.236                & 0.646                \\
\rowcolor[HTML]{EFEFEF} 
DHA (IS =   512, A1) & \textbf{0.652}       & {\ul \textbf{0.139}} & \textbf{0.661}       & 0.624                & 0.213                & 0.641                & {\ul \textbf{0.615}} & 0.235                & 0.647                \\
\rowcolor[HTML]{EFEFEF} 
DHA (IS =   64, A2) $\;$  & \textbf{0.659}       & {\ul \textbf{0.139}} & {\ul \textbf{0.667}} & {\ul \textbf{0.629}} & {\ul \textbf{0.212}} & {\ul \textbf{0.647}} & {\ul \textbf{0.615}} & 0.234                & 0.653                \\
\rowcolor[HTML]{EFEFEF} 
DHA (IS =   256, A2) & \textbf{0.659}       & {\ul \textbf{0.139}} & {\ul \textbf{0.667}} & {\ul \textbf{0.629}} & {\ul \textbf{0.212}} & {\ul \textbf{0.648}} & {\ul \textbf{0.615}} & 0.234                & 0.654                \\
\rowcolor[HTML]{EFEFEF} 
DHA (IS =   512, A2) & {\ul \textbf{0.660}} & {\ul \textbf{0.139}} & {\ul \textbf{0.667}} & {\ul \textbf{0.629}} & {\ul \textbf{0.212}} & {\ul \textbf{0.647}} & {\ul \textbf{0.615}} & 0.234                & 0.654               
\\
\Xhline{0.1\arrayrulewidth}
\bottomrule
\end{tabular}
\end{adjustbox}
\vspace{0.2cm}
\caption{ \color{black} The results on the SUPPORT dataset. For C-Index and ROC, higher scores are better. For BS, lower is better. The best mean results are \textbf{\underline{underlined}} and the results that are close to the best one, by repeated k-fold cv t-test statistics in 95\% confident interval, are shown in \textbf{bold}. \label{tab:sup}}
\end{table}

\paragraph{Concordance Index (C-Index).}
C-Index is the probability that predicted survival durations for two instances have the same ordering as their actual survival times. Initially, C-Index was derived for proportional hazards framework by \citet{harrell1982evaluating} and later extended for non-proportional cases \citep{antolini2005time}. More recently, \citet{uno2011c} proposed to correct C-Index for censoring using IPCW which is what we employ in this paper:
\begin{equation}
    \mathrm{C}(t) = P ( S(t|\boldsymbol{x}_i) < S(t|\boldsymbol{x}_j) | t_i < t_j, t_i<t ).
\end{equation}
\paragraph{Brier Score (BS).} A perfect ranking can be obtained without assessing appropriate risk scores, which results in calibration problem. BS measures the model calibration by the expected square difference between the survival predictions and event indicators:
\begin{equation}
    \mathrm{BS}(t) = \EX\left[ (I_{t_i>t} - S(t|\boldsymbol{x}_i))^2 \right].
\end{equation}
BS is originally derived to evaluate the accuracy of weather forecasts by \citet{brier1950verification} and later extended to censored time-to-event datasets by \citet{graf1999assessment} which is what we employ in this paper. 
\paragraph{Area Under Receiver Operating Curve (ROC-AUC).} ROC-AUC quantifies the seperation of positive and negative instances where the positives are defined as the instances that experienced the event before time $t$: 
\begin{align}
\begin{split}
\mathrm{AUC}(t) = P(S(t|\mathbf{x}_i) \leq  S(t|\mathbf{x}_j)  | t_i \leq t,  t_j > t)
\end{split}\label{auc}
\end{align}
This definition also relates to the time-dependent C-Index derived by \citet{antolini2005time}, which is based on the sum of weighted AUC scores at different time steps. Similar to previous metrics, we adjust this measure for censoring to have an unbiased estimate \citep{hung2010estimation,kamarudin2017time}.

\begin{table*}[]
\center
\begin{adjustbox}{width=\textwidth}
\begin{tabular}{c|ccc|ccc|ccc}
\toprule
\multicolumn{10}{c}{$\;\;\;\;\;\;\;\;\;\;\;\;\;\;\;\;$METABRIC Dataset}                                                                                          \\ \midrule
                         & \multicolumn{3}{c}{25$^{\text{th}}$   Quantile}              & \multicolumn{3}{|c|}{50$^{\text{th}}$   Quantile}               & \multicolumn{3}{c}{75$^{\text{th}}$  Quantile}             
                         \\
                         \cline{2-10} 
\multirow{-2}{*}{Models} & C-Index        & BS             & ROC            & C-Index        & BS             & ROC            & C-Index        & BS             & ROC            \\ \hline
CoxPH                 & 0.629                 & 0.244                     & 0.640                 & 0.627                & 0.196                    & 0.649                & 0.633                 & 0.334                     & 0.684                 \\
DeepSurv              & 0.640                 & 0.122                     & 0.653                 & 0.636                & 0.197                    & 0.657                & 0.635                 & 0.227                     & 0.677                 \\
RSF                   & \textbf{0.702}        & \textbf{0.117}            & 0.718                 & \textbf{0.669}       & \textbf{0.192}           & \textbf{0.689}       & \textbf{0.639}        & 0.227                     & 0.675                 \\
DeepHit               & \textbf{0.703}        & {\ul \textbf{0.116}}      & \textbf{0.719}        & 0.653                & 0.194                    & 0.674                & 0.616                 & 0.230                     & 0.665                 \\
Cox-Time              & \textbf{0.703}        & \textbf{0.117}            & \textbf{0.720}        & 0.665                & \textbf{0.191}           & 0.688                & \textbf{0.638}        & \textbf{0.226}            & \textbf{0.681}        \\
DSM                   & \textbf{0.701}        & \textbf{0.118}            & 0.715                 & 0.667                & 0.209                    & 0.685                & \textbf{0.641}        & 0.261                     & \textbf{0.672}        \\
DCM                   & \textbf{0.698}        & 0.122                     & 0.714                 & 0.663                & 0.200                    & 0.682                & \textbf{0.637}        & 0.232                     & \textbf{0.673}        \\
\Xhline{0.1\arrayrulewidth}
\rowcolor[HTML]{EFEFEF} 
DHA (IS =   64, A1) $\;$  & \textbf{0.706}        & \textbf{0.117}            & \textbf{0.720}        & \textbf{0.670}       & \textbf{0.192}           & \textbf{0.690}       & {\ul \textbf{0.647}}  & \textbf{0.226}            & {\ul \textbf{0.686}}  \\
\rowcolor[HTML]{EFEFEF} 
DHA (IS =   256, A1) & \textbf{0.708}        & \textbf{0.117}            & \textbf{0.722}        & \textbf{0.670}       & \textbf{0.192}           & 0.688                & \textbf{0.643}        & 0.227                     & \textbf{0.683}        \\
\rowcolor[HTML]{EFEFEF} 
DHA (IS =   512, A1) & \textbf{0.706}        & \textbf{0.117}            & \textbf{0.721}        & 0.668                & 0.193                    & 0.688                & \textbf{0.643}        & 0.227                     & \textbf{0.684}        \\
\rowcolor[HTML]{EFEFEF} 
DHA (IS =   64, A2) $\;$  & \textbf{0.710}        & \textbf{0.117}            & \textbf{0.725}        & \textbf{0.672}       & \textbf{0.192}           & \textbf{0.692}       & 0.637                 & 0.227                     & \textbf{0.677}        \\
\rowcolor[HTML]{EFEFEF} 
DHA (IS =   256, A2) & {\ul \textbf{0.712}}  & {\ul \textbf{0.116}}      & {\ul \textbf{0.726}}  & \textbf{0.675}       & \textbf{0.191}           & \textbf{0.695}       & \textbf{0.639}                 & \textbf{0.225}            & \textbf{0.680}        \\
\rowcolor[HTML]{EFEFEF} 
DHA (IS =   512, A2) & \textbf{0.710}        & {\ul \textbf{0.116}}      & \textbf{0.725}        & {\ul \textbf{0.676}} & {\ul \textbf{0.190}}     & {\ul \textbf{0.696}} & \textbf{0.643}        & {\ul \textbf{0.224}}      & \textbf{0.684}       
                       
\\
\Xhline{0.1\arrayrulewidth}
\bottomrule
\end{tabular}
\end{adjustbox}
\vspace{0.2cm}
\caption{ \color{black} The results on the METABRIC dataset. For C-Index and ROC, higher scores are better. For BS, lower is better. The best mean results are \textbf{\underline{underlined}} and the results that are close to the best one, by repeated k-fold cv t-test statistics in 95\% confident interval, are shown in \textbf{bold}. \label{tab:met}}
\end{table*}

\subsection{Experimental Design}\label{sec:des}

We perform 2x5-fold cross-validation (cv) for our model and baselines. The random seeds are fixed, and each train-valid-test splits seen by the models are identical for all datasets within runs. We use t-test with corrected repeated k-fold cv test to correct our t-statistics for the correlation between splits \citep{bouckaert2004evaluating}:

\begin{equation}
    t^{l} = \frac{\mu^{l}}{\hat{\sigma}^{{l}^2}\sqrt{\frac{1}{kr} + \frac{n_{\text{te}}}{n_{\text{tr}}}}}.
\end{equation}

Here, $\mu^{l} = \frac{1}{kr}\sum_{i=1}^k\sum_{j=1}^r y^{l}_{ij} $, where $y^{l}_{ij}$ corresponds to performance difference between two models for $i^{\text{th}}$ fold and $j^{\text{th}}$ run on $l^{\text{th}}$ metric. $k$ and $r$ are defined as number of folds and runs, and $n_{\text{tr}}$ and $n_{\text{te}}$ are train and test sizes, respectively. Finally, $\hat{\sigma}^{{l}^2} = \frac{1}{kr -1} \sum_{i=1}^k\sum_{j=1}^r (y^{l}_{ij} - \mu^{l})$. The test statistic $t^{l}$ is distributed according to Student's t-distributions with $kr - 1$ degrees of freedom.

We report the mean results for 25$^{\text{th}}$, 50$^{\text{th}}$, and 75$^{\text{th}}$ quantiles for each metric, and highlight them with respect to the t-statistic described above.

Each baseline model has been fully tuned for each dataset using the validation set. The models are trained for 4000 epochs until convergence on each fold and run. To ensure a fair comparison, we perform early stopping on all models using their validation loss. We changed the IS size while using fixed hyperparameters in our model to study the effects of IS. We describe the hyperparameter optimization protocol in Appendix \ref{ap:hyp}. We use a single Nvidia GeForce RTX 20 series graphics card to carry-out our experiments. We refer reader to \citet{nagpal2022auton}, \citet{havakvpy36:online}, and \citet{sksurv} for baseline implementations. 

\section{Results and Discussion}\label{sec:res}

\begin{table*}[]
\center
\begin{adjustbox}{width=\textwidth}
\begin{tabular}{c|ccc|ccc|ccc}
\toprule
\multicolumn{10}{c}{$\;\;\;\;\;\;\;\;\;\;\;\;\;\;\;\;$FLCHAIN Dataset}                                                                                          \\ \midrule
                         & \multicolumn{3}{c}{25$^{\text{th}}$   Quantile}              & \multicolumn{3}{|c|}{50$^{\text{th}}$   Quantile}               & \multicolumn{3}{c}{75$^{\text{th}}$  Quantile}             
                         \\
                         \cline{2-10} 
\multirow{-2}{*}{Models} & C-Index        & BS             & ROC            & C-Index        & BS             & ROC            & C-Index        & BS             & ROC            \\ \hline
CoxPH                 & 0.789                & 0.103                & 0.800                & \textbf{0.793}       & \textbf{0.098}       & \textbf{0.816}       & \textbf{0.791}       & 0.168                & \textbf{0.826}       \\
DeepSurv              & 0.786                & \textbf{0.060}       & 0.797                & 0.790                & \textbf{0.100}       & 0.813                & 0.788                & \textbf{0.126}       & \textbf{0.823}       \\
RSF                   & {\ul \textbf{0.801}} & {\ul \textbf{0.058}} & {\ul \textbf{0.813}} & {\ul \textbf{0.796}} & {\ul \textbf{0.098}} & {\ul \textbf{0.819}} & {\ul \textbf{0.792}} & {\ul \textbf{0.124}} & {\ul \textbf{0.827}} \\
DeepHit               & 0.792                & 0.061                & 0.803                & \textbf{0.794}       & \textbf{0.101}       & \textbf{0.817}       & \textbf{0.790}       & \textbf{0.127}       & \textbf{0.825}       \\
Cox-Time              & \textbf{0.795}       & 0.066                & \textbf{0.807}       & {\ul \textbf{0.796}} & 0.120                & {\ul \textbf{0.819}} & {\ul \textbf{0.792}} & 0.165                & {\ul \textbf{0.827}} \\
DSM                   & \textbf{0.791}       & 0.061                & 0.803                & \textbf{0.793}       & 0.111                & \textbf{0.815}       & \textbf{0.790}       & 0.147                & \textbf{0.825}       \\
DCM                   & \textbf{0.793}       & \textbf{0.059}       & \textbf{0.805}       & 0.785                & \textbf{0.101}       & 0.806                & \textbf{0.780}       & 0.128                & 0.813                \\
\rowcolor[HTML]{EFEFEF} 
\Xhline{0.1\arrayrulewidth}
DHA (IS =   64, A1) $\;$  & \textbf{0.793}       & 0.063                & 0.804                & 0.792                & 0.109                & 0.814                & \textbf{0.790}       & 0.143                & 0.822                \\
\rowcolor[HTML]{EFEFEF} 
DHA (IS =   256, A1) & \textbf{0.793}       & 0.063                & 0.804                & \textbf{0.793}       & 0.110                & \textbf{0.815}       & 0.789                & 0.145                & 0.822                \\
\rowcolor[HTML]{EFEFEF} 
DHA (IS =   512, A1) & \textbf{0.793}       & 0.063                & 0.803                & \textbf{0.793}       & 0.109                & 0.814                & 0.789                & 0.143                & 0.822                \\
\rowcolor[HTML]{EFEFEF} 
DHA (IS =   64, A2) $\;$  & \textbf{0.799}       & 0.061                & \textbf{0.810}       & \textbf{0.795}       & 0.105                & \textbf{0.817}       & \textbf{0.790}       & 0.138                & 0.823                \\
\rowcolor[HTML]{EFEFEF} 
DHA (IS =   256, A2) & \textbf{0.799}       & 0.062                & \textbf{0.811}       & \textbf{0.794}       & 0.106                & \textbf{0.816}       & 0.788                & 0.140                & 0.821                \\
\rowcolor[HTML]{EFEFEF} 
DHA (IS =   512, A2) & \textbf{0.800}       & 0.061                & \textbf{0.812}       & \textbf{0.795}       & 0.105                & \textbf{0.817}       & 0.789                & 0.139                & 0.822               

\\
\Xhline{0.1\arrayrulewidth}
\bottomrule
\end{tabular}
\end{adjustbox}
\vspace{0.2cm}
\caption{ \color{black} The results on the FLCHAIN dataset. For C-Index and ROC, higher scores are better. For BS, lower is better. The best mean results are \textbf{\underline{underlined}} and the results that are close to the best one, by repeated k-fold cv t-test statistics in 95\% confident interval, are shown in \textbf{bold}. \label{tab:fl}}
\end{table*}

For the SUPPORT dataset, our approach yields the best performance results over 25$^\text{th}$ and 50$^\text{th}$ event quantiles. For the 75$^\text{th}$ quantile RSF and Cox-Time are on par, while we yield the best results on C-Index.



For the METABRIC dataset, Cox-Time, RSF and our approach perform well across shorter and longer time-horizons with our approach having the best average results over shorter and longer horizons for different importance sampling sizes. 

For the FLCHAIN dataset, we see that RSF retains the best average result while our approach is comparable on ranking based metrics.

Finally, for CKD/AKI dataset, our approach retains the best mean results across most of the metrics over both shorter and longer time horizons while DeepHit being our closest competitor.


Overall, our approach consistently demonstrates better results in \textbf{29} out of 36 dataset-metric pairs with \textbf{\underline{21}} out of \textbf{29} being the best average, over different baseline models, including continuous and discrete state-of-the-art approaches. Our closest competitor is RSF, which demonstrates better results in \textbf{30} out of 36 metrics with \textbf{\underline{12}} out of 30 being the best average. 

To summarize our results, we emphasize several important points:  \textbf{(1)} For neural non-proportional hazard modeling having a separate embedding layer (A2) for time is more beneficial than concatenating and feeding everything to a shared neural network (A1), \textbf{(2)} despite being introduced much earlier,  when tuned carefully, RSF performs on par or better than the other models. \textbf{(3)} best-performing benchmark models differ between datasets and time-horizons, confirming the findings of \citet{lee2019dynamic}, \textbf{(4)} our model performs well consistently over different datasets and time horizons with minimal hyperparameter tuning. \textbf{(5)} parametric continuous-time models are more robust to hyperparameter choice while discrete-time models (e.g., DeepHit) have critical hyperparameters as also emphasized by \citet{sloma2021empirical}.\footnote{In particular, we found that DeepHit is sensitive to the number of output neurons (\textquoteleft num\_durations') and must be tuned carefully: too many results in the training of very few, and too few result in information loss.} \textbf{(6)} altering the IS size does not result in a significant change, which shows the robustness of our approach.

\begin{table*}[]
\center
\begin{adjustbox}{width=\textwidth}
\begin{tabular}{c|ccc|ccc|ccc}
\toprule
\multicolumn{10}{c}{$\;\;\;\;\;\;\;\;\;\;\;\;\;\;\;\;$CKD/AKI Dataset}                                                                                          \\ \midrule
                         & \multicolumn{3}{c}{25$^{\text{th}}$   Quantile}              & \multicolumn{3}{|c|}{50$^{\text{th}}$   Quantile}               & \multicolumn{3}{c}{75$^{\text{th}}$  Quantile}             
                         \\
                         \cline{2-10} 
\multirow{-2}{*}{Models} & C-Index        & BS             & ROC            & C-Index        & BS             & ROC            & C-Index        & BS             & ROC            \\ \hline
CoxPH                 & 0.598                & 0.089                & 0.612                & 0.581                & 0.167                & 0.614                & 0.575                & 0.225                & 0.649                \\
DeepSurv              & 0.619                & \textbf{0.088}       & 0.633                & 0.607                & 0.164                & 0.643                & 0.603                & \textbf{0.219}       & 0.673                \\
RSF                   & \textbf{0.639}       & {\ul \textbf{0.087}} & \textbf{0.655}       & \textbf{0.620}       & \textbf{0.163}       & \textbf{0.657}       & \textbf{0.608}       & {\ul \textbf{0.218}} & \textbf{0.686}       \\
DeepHit               & \textbf{0.642}       & {\ul \textbf{0.087}} & \textbf{0.658}       & \textbf{0.624}       & {\ul \textbf{0.162}} & {\ul \textbf{0.663}} & \textbf{0.610}       & {\ul \textbf{0.218}} & \textbf{0.686}       \\
Cox-Time              & 0.623                & \textbf{0.089}       & 0.635                & 0.613                & 0.165                & 0.649                & \textbf{0.605}       & 0.220                & 0.676                \\
DSM                   & \textbf{0.635}       & \textbf{0.088}       & 0.647                & \textbf{0.615}       & 0.182                & \textbf{0.658}       & 0.602                & 0.247                & {\ul \textbf{0.688}} \\
\Xhline{0.1\arrayrulewidth}
\rowcolor[HTML]{EFEFEF} 
DHA (IS =   64, A1) $\;$  & \textbf{0.637}       & {\ul \textbf{0.087}} & 0.650                & \textbf{0.617}       & \textbf{0.163}       & 0.653                & \textbf{0.605}       & 0.220                & \textbf{0.685}       \\
\rowcolor[HTML]{EFEFEF} 
DHA (IS =   256, A1) & \textbf{0.638}       & \textbf{0.088}       & 0.652                & \textbf{0.618}       & \textbf{0.163}       & 0.654                & \textbf{0.605}       & 0.220                & 0.682                \\
\rowcolor[HTML]{EFEFEF} 
DHA (IS =   512, A1) & \textbf{0.637}       & \textbf{0.088}       & 0.649                & \textbf{0.617}       & \textbf{0.163}       & 0.653                & \textbf{0.605}       & 0.221                & \textbf{0.683}       \\
\rowcolor[HTML]{EFEFEF} 
DHA (IS =   64, A2) $\;$  & \textbf{0.644}       & {\ul \textbf{0.087}} & {\ul \textbf{0.660}} & {\ul \textbf{0.625}} & {\ul \textbf{0.162}} & \textbf{0.661}       & \textbf{0.610}       & {\ul \textbf{0.218}} & \textbf{0.685}       \\
\rowcolor[HTML]{EFEFEF} 
DHA (IS =   256, A2) & {\ul \textbf{0.646}} & {\ul \textbf{0.087}} & {\ul \textbf{0.660}} & {\ul \textbf{0.625}} & {\ul \textbf{0.162}} & \textbf{0.659}       & {\ul \textbf{0.611}} & {\ul \textbf{0.218}} & \textbf{0.683}       \\
\rowcolor[HTML]{EFEFEF} 
DHA (IS =   512, A2) & \textbf{0.642}       & {\ul \textbf{0.087}} & \textbf{0.657}       & \textbf{0.623}       & 0.164                & \textbf{0.659}       & \textbf{0.608}       & {\ul \textbf{0.218}} & \textbf{0.683}          
\\
\Xhline{0.1\arrayrulewidth}
\bottomrule
\end{tabular}
\end{adjustbox}
\vspace{0.2cm}
\caption{ \color{black} The results on the CKD/AKI dataset. For C-Index and ROC, higher scores are better. For BS, lower is better. The best mean results are \textbf{\underline{underlined}} and the results that are close to the best one, by repeated k-fold cv t-test statistics in 95\% confident interval, are shown in \textbf{bold}. We found DCM to be unstable on this dataset and did not include it.\label{tab:ckd}}
\end{table*}

\section{Limitations and Future Work}

We consider a number of limitations for this work to address in the future:

\paragraph{Competing Risk Scenarios.} We consider extending our approach for competing-risk scenarios in which various events may lead to failure. In particular, modifying our architecture to accommodate competing risks utilizing a common covariate layer and sub-networks for each competing event, and adjusting the likelihood may enable information flow from various risks an instance confronts. This approach is similar to DeepHit  \citep{lee2018deephit}.
\paragraph{Temporal Data.}  Certain time-to-event data, such as vital signs and electronic health records, can consist of time series. In such cases, leveraging the temporal structure of data is important. Similar to \citet{nagpal2021deep}, we consider altering the network architecture to account for the temporality of the clinical data using recurrent neural networks (RNNs). 
\paragraph{Different Modalities.} Another potential direction for this work includes using different data modalities to perform survival analysis, such as medical imaging. We consider altering the network to incorporate image data using Convolutional Neural Networks (CNNs) like \citet{zhu2016deep}.
\paragraph{Small Datasets.} DHA is a deep learning model. We acknowledge that deep learning models require large datasets and may demonstrate inferior performance compared to non-parametric approaches, like RSF, when dealing with small datasets.
\section{Conclusion}





In this work, we demonstrate that there are a number of undesirable characteristics in existing state of the art survival models. In particular, discrete time models and mixture density models are very sensitive to hyperparameters such as number of bins and mixtures. Each of these choices requires extensive tuning by practitioners to achieve optimal performance. We introduce a method which is free of such hyperparameters and exhibits all the desirable properties in existing state of the art methods such as unbiased exact log-likelihood maximization, flexibility in density estimation and continuous-time. We train our model with default parameters on all datasets and it is able to match or outperform existing state of the art methods. We believe this model will ease the burden on practitioners for fitting new datasets. 

\section{Acknowledgements}
Funding for this work is from NHLBI award R01HL148248.



\appendix
\section*{Hyperparameters}\label{ap:hyp}

 All models have an equal training length of 4000 epochs. We pick the best-performing model with respect to their validation loss. The hyper-parameter spaces of each benchmark model are listed below.

\paragraph{CoxPH.}

    \begin{itemize}
        \item[] 
            \textquoteleft alpha': [0, 1e-3, 1e-2, 1e-1],
    \end{itemize}

\paragraph{DeepSurv.}

    \begin{itemize}
        \item[] 
        
            \textquoteleft lr' : [5e-4, 1e-3],

            \textquoteleft batch\_size': [256, 512, 1024],

            \textquoteleft weight\_decay': [0, 1e-8, 1e-6, 1e-3, 1e-1],

            \textquoteleft nodes\_':[128, 256, 512],

            \textquoteleft layers\_': [2, 3],

            \textquoteleft dropout': [0, 1e-1, 2e-1, 4e-1, 5e-1],

    \end{itemize}

\paragraph{RSF.}

    \begin{itemize}
        \item[] 
                \textquoteleft max\_depth' : [None, 5],

                \textquoteleft n\_estimators' : [50, 100, 150, 200, 150],
                
                \textquoteleft max\_features' : [50, 75, sqrt(d), d//2, d],

                \textquoteleft min\_samples\_split' : [10, 150, 200, 250],
                
    \end{itemize}

\textquoteleft max\_depth':None means that the expansion continues until all leaves are pure.

\paragraph{DSM.}

    \begin{itemize}
        \item[] 
        
         \textquoteleft k \textquoteright : [3, 4, 6],
         
         \textquoteleft distribution' : [\textquoteleft Weibull', \textquoteleft LogNormal'],
        
        \textquoteleft learning\_rate' : [1e-4, 5e-4, 1e-3],

        \textquoteleft nodes\_' : [48, 64, 96, 256],

        \textquoteleft hidden\_layers\_': [1, 2, 3],
        
        \textquoteleft discount': [1/3, 3/4, 1],
        
        \textquoteleft batch\_size': [128, 256],

    \end{itemize}

\paragraph{DCM.}

    \begin{itemize}
        \item[] 
        
            \textquoteleft k' : [3, 4, 6],
            
            \textquoteleft nodes\_' : [48, 64, 96, 256],
            
            \textquoteleft hidden\_layers\_': [1, 2, 3],
            
            \textquoteleft batch\_size': [128, 256],
            
            \textquoteleft use\_activation': [True, False],

    \end{itemize}
    
\paragraph{Deep-Hit.}

    \begin{itemize}
        \item[] 
        
            \textquoteleft lr' : [5e-4, 1e-3],

            \textquoteleft batch\_size': [256, 512, 1024],
            
            \textquoteleft weight\_decay': [0, 1e-8, 1e-6, 1e-3, 1e-1],

            \textquoteleft nodes\_':[128, 256, 512],

            \textquoteleft hidden\_layers\_': [2, 3],

            \textquoteleft dropout': [0, 1e-1, 2e-1, 4e-1, 5e-1],

            \textquoteleft alpha': [1e-1, 2e-1, 4e-1, 8e-1, 1],
  
            \textquoteleft sigma' : [1e-1, 2.5e-1, 4e-1, 8e-1, 1, 2, 10],

            \textquoteleft num\_durations' : [ 10, 50, 100],

    \end{itemize}

\paragraph{Cox-Time.}

    \begin{itemize}
        \item[] 
        
            \textquoteleft lr' : [5e-4, 1e-3],

            \textquoteleft batch\_size': [256, 512, 1024],
            
            \textquoteleft weight\_decay': [0, 1e-8, 1e-6, 1e-3, 1e-1],

            \textquoteleft nodes\_':[128, 256, 512],

            \textquoteleft hidden\_layers\_': [1,2],

            \textquoteleft dropout': [0, 1e-1, 2e-1, 4e-1, 5e-1],

            \textquoteleft lambda': [0, 1e-3, 1e-2, 1e-1],
  
            \textquoteleft log\_duration' : [True, False],

    \end{itemize}

\paragraph{Ours.}

    \begin{itemize}
        \item[] 
        
            \textquoteleft lr' : 2e-3,

            \textquoteleft batch\_size': 256,

            \textquoteleft imps\_size': [64, 256, 512],

            \textquoteleft architecture': [\textquoteleft A1', \textquoteleft A2'],

            \textquoteleft layer\_norm' : True,
            
            \textquoteleft weight\_decay': 1e-5,

            \textquoteleft nodes\_': 400

            \textquoteleft layers\_': 2,

            \textquoteleft dropout': 4e-1,

            \textquoteleft act': selu,

    \end{itemize}

\end{document}